\documentclass{article}
\usepackage{spconf,amsmath,graphicx}

\title{Constructing Long Short-Term Memory based Deep Recurrent Neural Networks for Large Vocabulary Speech Recognition}
\name{Xiangang Li, Xihong Wu}
\address{Speech and Hearing Research Center, \\
Key Laboratory of Machine Perception (Ministry of Education), \\
Peking University, Beijing, 100871\\
{\small \tt \{lixg, wxh\}@cis.pku.edu.cn}}
\begin{document}
\ninept
\maketitle
\begin{abstract}
Long short-term memory (LSTM) based acoustic modeling methods have recently been shown to give state-of-the-art performance on some speech recognition tasks. To achieve a further performance improvement, in this research, deep extensions on LSTM are investigated considering that deep hierarchical model has turned out to be more efficient than a shallow one. Motivated by previous research on constructing deep recurrent neural networks (RNNs), alternative deep LSTM architectures are proposed and empirically evaluated on a large vocabulary conversational telephone speech recognition task. Meanwhile, regarding to multi-GPU devices, the training process for LSTM networks is introduced and discussed. Experimental results demonstrate that the deep LSTM networks benefit from the depth and yield the state-of-the-art performance on this task.
\end{abstract}
\section{Introduction}
\label{sec:intro}
Recently, the context dependent (CD) deep neural network (DNN) hidden Markov model (HMM) (CD-DNN-HMM) has become the dominant framework for acoustic modeling in speech recognition (e.g. \cite{AmDBN}\cite{DnnForAm}\cite{CdDnn1}\cite{CdDnn2}). However, given that speech is an inherently dynamic process, some researchers pointed out that recurrent neural networks (RNNs) can be considered as alternative models for acoustic modeling \cite{SRWDRNN}. The cyclic connections in RNNs exploit a self-learnt amount of temporal context, which makes RNNs better suited for sequence modeling tasks. Unfortunately, in practice, conventional RNNs are hard to be trained properly due to the vanishing gradient and exploding gradient problems as described in \cite{LRNN}. To address these problems, literature \cite{LSTM1} proposed an elegant RNN architecture, called as long short-term memory (LSTM).

%DNNs operate the input sequence on a fixed-size sliding window, making them unsuited to handle different speaking rates and longer dependencies. RNNs have cyclic connections which allow them to exploit dynamically changing contextual window over the input sequence history. This capability makes RNNs better suited for sequence modeling tasks.

%DNNs operate the input sequence on a fixed-size sliding window, making them unsuited to handle different speaking rates and longer dependencies, while RNNs have cyclic connections which allow them to exploit dynamically changing contextual window over the input sequence history, and this makes RNNs a natural choice for sequence modeling tasks.

LSTMs and conventional RNNs have been successfully used for many sequence labeling and sequence prediction tasks. In language modeling, RNNs were used as generative models over word sequences,
and remarkable improvements were achieved \cite{RNNLM} over the standard n-gram models. For handwriting recognition, LSTM networks have been applied for a long time \cite{handwriting}, in which, the bidirectional LSTM (BLSTM) networks trained with connectionist temporal classification (CTC)\cite{CTC} has been demonstrated performing better than the HMM-based system. In speech synthesis, the BLSTM network has also been applied and a notable improvement was obtained \cite{LSTM.TTS}. For language identification, LSTM based approach was proposed in \cite{LSTM.LID} to compared with i-vector and DNN systems, and better performance was achieved. Recently, LSTM networks have also been introduced on phoneme recognition task \cite{SRWDRNN}, robust speech recognition task \cite{RobustASR.LSTM}, and large vocabulary speech recognition task \cite{HSRWDBLSTM}\cite{LSTMRNNAM}\cite{LSTMLVSP}, and shown state-of-the-art performances. Subsequently, the sequence discriminative training of LSTM networks is investigated in \cite{SeqDis.LSTM}, and a significant gain was obtained.

%Recently, LSTM networks have also been proposed and shown to give the state-of-the-art performance on phoneme recognition \cite{SRWDRNN} and large vocabulary speech recognition tasks \cite{HSRWDBLSTM}\cite{LSTMRNNAM}\cite{LSTMLVSP}. which showed that, the LSTM based models can give the state-of-the-art performance. which give the state-of-the-art performance.

In the researches of acoustic modeling, depth for feed-forward neural networks can lead to more expressive models. LSTMs and conventional RNNs are inherently deep in time, for they can be expressed as a composition of multiple nonlinear layers when unfolded in time. This paper explores the depth of LSTMs, which is defined as the depth in space. Based on earlier researches on constructing deep RNNs \cite{HTCDRNN}, in this work, possible approaches are explored to extend LSTM networks into deep ones, and various deep LSTM networks are empirically evaluated and compared on a large vocabulary Mandarin Chinese conversational telephone speech recognition task. Although lots of attentions have been attracted to the deep LSTM networks, this paper summaries the approaches of constructing deep LSTM networks from different perspectives, and suggests alternative architectures that can yield comparable performance.

%deep feed-forward neural networks have been shown to be more expressive models. \cite{SRWDRNN} earlier proposed a way of building deep LSTM by stacking multiple LSTM layers. \cite{LSTMRNNAM} proposed an alternative LSTM architecture, in which a projection layer is connected with the cell output units. \cite{SCUTHFEFDNN} use the high-level features extracted from deep neural network as the input for RNNs. What's more, in \cite{HTCDRNN}, many ways are explored to construct deep RNN. in this paper, we investigate possible approaches to extend LSTMs into deep LSTMs, and then applied these approaches for acoustic modeling in a large vocabulary speech recognition task.

%The rest of this paper is organized as follows. Section 2 describes the conventional LSTM architecture and the approaches of constructing LSTM based deep networks. Section 3 presents the implementation of LSTM network training with multi-GPU devices. Experiments and results are presented in Section 4. Finally, Section 5 concludes the paper.

\section{Constructing LSTM based deep RNNs}
\subsection{The conventional LSTM architecture}

Given an input sequence $x=(x_1,x_2,\ldots,x_T)$, a conventional RNN computes the hidden vector sequence $h=(h_1,h_2,\ldots,h_T)$ and output vector sequence $y=(y_1,y_2,\ldots, y_T)$ from $t=1$ to $T$ as follows:
\begin{align}\label{RNN}
    &h_t=\mathcal{H}(W_{xh}x_{t}+W_{hh}h_{t-1}+b_{h})\\
    &y_t=W_{hy}h_{t}+b_{y}
\end{align}
where, the $W$ denotes weight matrices, the $b$ denotes bias vectors and $\mathcal{H}(\cdot)$ is the recurrent hidden layer function.

\begin{figure}[htb]
\centerline{\includegraphics[width=65mm]{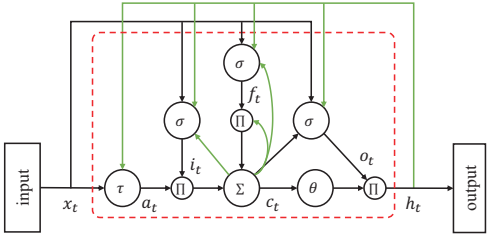}}
\caption{\label{Figure0} {The architecture of a LSTM network with one memory block, where green lines are time-delayed connections.}}
\label{spprod}
\end{figure}

In the LSTM architecture, the recurrent hidden layer consists of a set of recurrently connected subnets known as ``memory blocks''. Each memory block contains one or more self-connected memory cells and three multiplicative gates to control the flow of information. In each LSTM cell, the flow of information into and out of the cell is guarded by the learned input and output gates.
Later, in order to provide a way for the cells to reset themselves, the forget gate was added \cite{LSTM.Forget.Gate}. In addition, the modern LSTM architecture contains peephole weights connecting the gates to the memory cell, which improve the LSTM's ability to learn tasks that require precise timing and counting of the internal states \cite{LSTM.Peephole.Weights}. As illustrated in Fig.~\ref{Figure0}, the recurrent hidden layer function $\mathcal{H}$ for this version of LSTM networks is implemented as following:

%In addition, there are peephole weights connecting the gates to the memory cell, which improve the LSTM's ability to learn precise timing and counting of the internal states \cite{LSTM.Peephole.Weights}. As illustrated in Fig.~\ref{Figure0}, the recurrent hidden layer function $\mathcal{H}$ for this version of LSTM networks is implemented as following:
\begin{align}\label{LSTM}
    &i_t=\sigma(W_{xi}x_t+W_{hi}h_{t-1}+W_{ci}c_{t-1}+b_{i})\\
    &f_t=\sigma(W_{xf}x_t+W_{hf}h_{t-1}+W_{cf}c_{t-1}+b_{f})\\
    &a_t=\tau(W_{xc}x_t+W_{hc}h_{t-1}+b_{c})\\
    &c_t=f_{t}c_{t-1}+i_{t}a_{t}\\
    &o_t=\sigma(W_{xo}x_t+W_{ho}h_{t-1}+W_{co}c_{t}+b_{o})\\
    &h_t=o_{t}\theta(c_t)
\end{align}
Where, $\sigma$ is the logistic sigmoid function, and $i$, $f$, $o$, $a$ and $c$ are respectively the input gate, forget gate, output gate, cell input activation, and cell state vectors, and all of which are the same size as the hidden vector $h$. $W_{ci}$, $W_{cf}$, $W_{co}$ are diagonal weight matrices for peephole connections. $\tau$ and $\theta$ are the cell input and cell output non-linear activation functions, generally in this paper $tanh$.

\subsection{Deep LSTM networks}
A number of theoretical results support that a deep, hierarchical model can be more efficient at representing some functions than a shallow one \cite{LearningDeepArchAI}. This paper is focused on constructing deep LSTM networks.
%Specifically for the acoustic modeling in speech recognition, as with deep feed-forward neural network, the deep LSTMs have been successfully used [].

\begin{figure}[htb]
\centerline{\includegraphics[width=75mm]{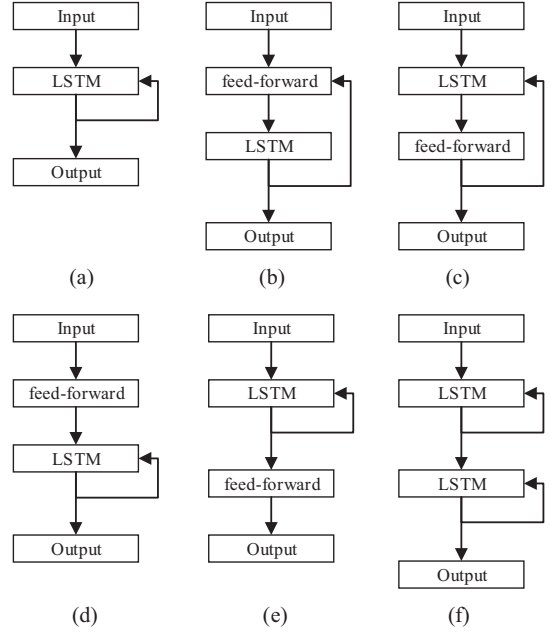}}
\caption{\label{Figure1} {Illustrations of different strategies for constructing LSTM based deep RNNs. (a) a conventional LSTM; (b) a LSTM with input projection; (c) a LSTM with output projection. (d) a LSTM with deep input-to-hidden function; (e) a LSTM with deep hidden-to-output function; (f) stacked LSTMs}}
\label{spprod}
\end{figure}

In \cite{HTCDRNN}, the architecture of conventional RNNs is carefully analyzed, and from three points, an RNN can be deepened: (1) input-to-hidden function, (2) hidden-to-hidden transition and (3) hidden-to-output function. In this paper, from these three points and the stacked LSTMs, several novel architectures to extend LSTM networks to deep ones are introduced as follows. For convenience, a simplified illustration of the LSTM is shown in Fig.~\ref{Figure1}(a) firstly.

%Novel architectures were constructed from these three points to increase the depth in space. %In order to introduce the proposed architecture, a simplified illustration of the LSTM network is shown in Fig.~\ref{Figure1}(a). %For the deep LSTM networks, we attempt to propose some novel architectures from these three points to increase the depth in space. In order to introduce the proposed architecture, a simplified illustration of the LSTM network is shown in Fig.~\ref{Figure1}(a).

\subsubsection{Deep hidden-to-hidden transition}
%However, we intend to add more hidden layers for the hidden-to-hidden transition in LSTM, and these layers can added before or after the LSTM memory blocks. and these can be realized by the architectures illustrated in Fig.~\ref{Figure1}(b) and Fig.~\ref{Figure1}(c).

In \cite{HTCDRNN}, an RNN with deep transition is discussed for increasing the depth of the hidden-to-hidden transition. Thus, two architectures can be obtained, as illustrated in Fig.~\ref{Figure1}(b) and Fig.~\ref{Figure1}(c). In details, in the architecture shown in Fig.~\ref{Figure1}(b), a multiple layer transformation is added before the cell input activation,
and which means that the calculation of $a_t$ in equation (5) is changed as:
\begin{align}\label{LSTM-IP}
    &a_{0,t}=\phi_{0}(W_{0,x}x_t+W_{0,h}h_{t-1}+b_{0})\\
    &a_t=\phi_{L}(W_{L}\phi_{L-1}(\ldots\phi_{1}(W_{1}a_{0,t}+b_{1}))+b_{L})
\end{align}
Where, $\phi$ is the activation function. In this paper, this architecture is called as the \emph{LSTM with input projection layer} (LSTM-IP for short).

Another architecture, shown in Fig.~\ref{Figure1}(c), has a separate hidden layer after the LSTM memory blocks, and the output activation $h_t$ is changed as $p_t$
\begin{equation}\label{RNN2}
    p_t=\phi_{L}(W_{L}\phi_{L-1}(\ldots\phi_{0}(W_{0}h_{t} + b_{0}))+b_{L})
\end{equation}
%Where, both non-linear and linear activation functions can be used as $\phi$.
We call this architecture as the \emph{LSTM with output projection layer} (LSTM-OP for short). However, this architecture is proposed earlier in literature \cite{LSTMRNNAM} to address the computation complexity of learning a LSTM network, in which it is called as the LSTM projected. From the perspective of this paper, this architecture is considered as a way to increase the depth of the hidden-to-hidden transition, although it may further be beneficial in tackling computation complexity issue.

It should be noticed that, in the LSTM-OP architecture, linear activation units can be used in projection layer, just like literature \cite{LSTMRNNAM} suggested. By contrast, there must be a non-linear activation (e.g. $tanh$) units used in the projection layer in the LSTM-IP.

\subsubsection{Deep input-to-hidden function}
A typical way to make the input-to-hidden function deep is using higher-level representations of DNNs as the input for RNNs. Literature \cite{SCUTHFEFDNN} reported that a better phoneme recognition performance could be achieved by applying this strategy for RNNs. %, although they did not jointly train the deep input-to-hidden function and the RNNs. 
All the previous studies are based on conventional RNNs, and in this research, this method is adopted for constructing deep LSTM networks as illustrated in Fig.~\ref{Figure1}(d), and applied to a large vocabulary speech recognition task.

%, and apply it to a large vocabulary speech recognition task. Previous work has shown that such high-level features extracted from DNNs tend to better disentangle the underlying factors of variation than the original input and helps overcome the problem of over-fitting in sequence classification tasks for the training of conventional RNNs. \cite{SCUTHFEFDNN} reported that a better phoneme recognition performance could be achieved by employing this deep input-to-hidden function, although they did not jointly train the deep input-to-hidden function and the RNNs. \cite{SCUTHFEFDNN} reported that a better phoneme recognition performance could be achieved by using high-level features extracted from DNNs as inputs for RNNs. %, although they did not jointly train the deep input-to-hidden function and the RNNs. %However, about this strategy, all the previous work was based on conventional RNNs. We tend to evaluate this strategy on LSTMs as illustrated in Fig.~\ref{Figure1}(d), and apply it to a large vocabulary speech recognition task. %However, when it comes to this strategy, all the previous studies were based on conventional RNNs. We intend to evaluate this strategy on LSTMs as illustrated in Fig.~\ref{Figure1}(d), and apply it to a large vocabulary speech recognition task.

\subsubsection{Deep hidden-to-output function}
It was discussed that a deep hidden-to-output function can be useful to disentangle the factors of variations in the hidden state \cite{HTCDRNN}. Based on this view, we construct a deep LSTM network shown in Fig.~\ref{Figure1}(e) by adding some intermediate layers between the output of the LSTM and the softmax layer.

\subsubsection{Stack of LSTMs}
Perhaps, the most straight-forward way to construct the deep LSTM network is to stack multiple LSTM layers on top of each other. %The output sequence of the lower LSTM layer forms the input sequence for the upper LSTM layer.
Specifically, output $h_t$ from the lower LSTM layer, is the input $x_t$ of the upper LSTM layer. This stacked LSTM networks can combine the multiple levels or representations with flexible use of long range context, and was introduced for acoustic modeling in speech recognition in \cite{SRWDRNN}, which showed that a significant performance improvement can be obtained compared with the shallow one.

\section{GPU Implementation}
\begin{figure}[htb]
\centerline{\includegraphics[width=80mm]{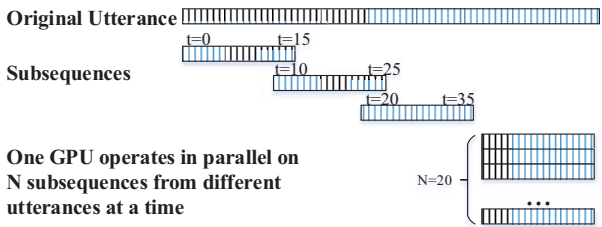}}
\caption{\label{Figure2} {Illustration of GPU implementation.}}
\label{spprod}
\end{figure}

We implement the LSTM network training on multi-GPU devices. In the training procedure, the truncated back-propagation though time (BPTT) learning algorithm \cite{BPTT} is adopted. Each sentence in the training set is split into subsequences with equal length $T_{bptt}$ (e.g. 15 frames). As illustrated in Fig.~\ref{Figure2}, two adjacent subsequences have overlapping frames $T_{overlap}$ (e.g. 5 frames). The gradients are computed for each subsequence and back-propagated to its start.
For computational efficiency, one GPU operates in parallel on $N$ (e.g. 20) subsequences from different utterances at a time. After the GPU has updated the parameters in the LSTM networks, it continues with the next $N$ subsequences in these utterances.%, preserving the LSTM state, or starts new utterances with reset state (a zero vector).
Besides, in order to train these networks on multi-GPU devices, asynchronous stochastic gradient descent (ASGD) \cite{ASGD1}\cite{ASGD2} is adopted.

In our experiments, it took us about two days to train a shallow conventional LSTM network having 750 cells with four GPU devices on a 150-hour speech corpus, where training a LSTM layer took around two to five times as much time as the training for a full-connection feed-forward hidden layer.

\section{Experiments}

We evaluate these LSTM networks on a large vocabulary speech recognition task - the HKUST Mandarin Chinese conversational telephone speech recognition \cite{HKUST}. The corpus (LDC2005S15, LDC2005T32) is collected and transcribed by Hong Kong University of Science and Technology (HKUST), which contains 150-hour speech, and 873 calls in the training set and 24 calls in the development set, respectively. In our experiments, around 10-hour speech was randomly selected from the training set, used as the validate set for network training, and the original development set in the corpus was used as speech recognition test set, which is not used in the training or the hyper-parameters determination procedures.

\subsection{Experimental setup}
The speech in the dataset is represented with 25ms frames of Mel-scale log-filterbank coefficients (including the energy value), along with their first and second temporal derivatives. In the experiments, the feed-forward DNNs used the concatenated features, which were produced by concatenating the current frame with 5 frames in its left and right context. However, for the inputs of LSTM networks, only current features (no context) were used.

%The speech in the dataset is processed with the standard short-time Fourier transform with a 25-ms Hamming window and with a fixed 10-ms frame rate. Raw speech features are generated subsequently using filterbank analysis, along with their first and second temporal derivatives. In the experiments, the feed-forward DNNs and LSTM networks with deep input-to-hidden function used the concatenated features, which were produced by concatenating the current frame with 5 frames in its left and right context. However, for the inputs of other LSTM networks, only current features were used (no context features were used).

A trigram language model was used in all the experiments, which was estimated using all the transcriptions of the acoustic model training set. We use a hybrid approach \cite{CdDnn2}\cite{LSTMRNNAM} for acoustic modeling with LSTM networks or DNNs, in which the neural networks' outputs are converted as pseudo likelihood as the state output probability in the HMM framework. All the networks were trained based on the alignments generated by a well-trained GMM-HMM systems with 3304 tied context dependent HMM states (realignments by DNNs were not performed), and only the cross-entropy objective function was used for all networks.

For network training, the learning rate was decreased exponentially. We tried to set the initial and final learning rates specific to a network architecture for stable convergence of each network. In the experiments, the initial learning rates ranged from 0.0005 to 0.002, and each final learning rate was always set as one-tenth of the corresponding initial one. In the training procedure of LSTM networks, the strategy introduced in \cite{RNN.difficulty} was applied to scale down the gradients. Besides, since the information from the future frames helps making LSTM networks better decisions for current frame, we also delayed the output HMM state labels by 3 frames.

\subsection{Experimental results}
Firstly, the baseline performance is summarized in Table~\ref{table0}.
For training the Subspace GMM \cite{SubspaceGMM}, KALDI toolkit \cite{KALDI} was used.
All the DNNs in the experiments had 4 hidden layers. Each layer in the ``ReLU DNN'' model had 2000 ReLU units \cite{ReLU}.
Each layer in the ``PNorm DNN'' model had 800 pnorm units \cite{Pnorm}, where the hyper-parameter $p$ is set to 2,
and the group size is set to 8. 
The ``Conv DNN'' model had two convolutional layers (along with max-pooling) and three ReLU layers. 
It can be found out that,
the character error rates (CER) of baseline GMM-HMM and DNN-HMM are comparable with those reported in \cite{HKUST.Result1}\cite{HKUST.Result2}\cite{HKUST.Result3}.

%The ``ReLU DNN'' model had 4 hidden layers with 2000 nodes each, and used the rectifier linear activation \cite{ReLU}. The ``PNorm DNN'' model had 4 hidden layers with 800 nodes each, and used the pnorm activation \cite{Pnorm} with $p=2$, $group size=8$. It can be found out that, the character error rates (CER) of baseline GMM-HMM and DNN-HMM are comparable with \cite{HKUST.Result1}\cite{HKUST.Result2}\cite{HKUST.Result3}.

\begin{table} [t,h]
\caption{\label{table0} {Speech recognition results of baseline systems on the HKUST Mandarin Chinese conversational telephone speech recognition task.}}
\vspace{2mm}
\centerline{
\begin{tabular}{|l|c|}
  \hline
  Model Descriptions & CER(\%) \\
  \hline \hline
  GMM & 48.68 \\
  \hline
  Subspace GMM & 44.29 \\
  \hline
  ReLU DNN & 38.42 \\
  \hline
  PNorm DNN & 38.01 \\
  \hline
  Conv DNN & 37.13 \\
  \hline
\end{tabular}}
\end{table}

Experiments were conducted to evaluate these deep LSTM networks shown in Fig.~{\ref{Figure1}}.
In the training procedure of these LSTM networks, the $T_{bptt}$ was fixed on 15, $T_{overlap}$ was fixed on 5.
Four GPUs were used, and each GPU operated in parallel on $20$ subsequences at a time.
%In the experiments, training a LSTM layer took around two to five times as much time as the training for a full-connection feed-forward hidden layer.

%For the LSTM-IP network, we added only one 1500 $tanh$ units hidden layer and had 750 LSTM cells.

The LSTM-IP network in the experiment had 750 LSTM cells and a non-linear activation projection layer with 2000 $tanh$ units.
The LSTM-OP network in the experiment had 2000 LSTM cells and a linear activation projection layer with 750 nodes.

In order to construct a LSTM network with deep input-to-hidden function, we constructed a LSTM network by putting a LSTM network on three feed-forward intermediate layers, and each feed-forward layer had 2000 ReLU units. This network is indicated as ``3-layer ReLU + LSTM'' in Table~\ref{table1}.
Similarly, we trained a model indicated as ``2-layer Conv + 2-layer ReLU + LSTM''. 
For the deep hidden-to-output function, a LSTM network, indicated as ``LSTM + 3-layer ReLU'' in Table~\ref{table1}, was constructed by adding three feed-forward intermediate hidden layers on top of the LSTM layer, and each feed-forward hidden layer had 2000 ReLU units.
The stacked LSTMs network was also evaluated, in which, three conventional LSTMs were stacked, and each layer had 750 LSTM cells. These three networks were trained using the discriminative pre-training algorithm \cite{DiscriminativePretraining}.
Concretely, in the training procedure of ``3-layer ReLU + LSTM'', three ReLU hidden layers were firstly pre-trained, and then the original output softmax layer was replaced by a new random initialized LSTM layer along with a new output softmax layer. Finally, the whole network was jointly optimized.
\begin{table} [t,h]
\caption{\label{table1} {Speech recognition results of different strategies of constructing deep LSTM networks.}}
\vspace{2mm}
\centerline{
\begin{tabular}{|l|c|}
  \hline
  Model Descriptions & CER(\%) \\
  \hline \hline
  LSTM & 40.28 \\
  \hline
  LSTM-IP & 39.09 \\
  \hline
  LSTM-OP & 35.92 \\
  \hline
  3-layer ReLU + LSTM & 37.31 \\
  \hline
  2-layer Conv + 2-layer ReLU + LSTM & 36.66 \\ 
  \hline
  LSTM + 3-layer ReLU & 37.16 \\
  \hline
  Stack of LSTM (3-layer) & \bf{35.91} \\
  \hline
\end{tabular}}
\end{table}

Comparing these results listed in Table~\ref{table1} with the baseline, the performance of 1-layer conventional LSTM network is even worse than the feed-forward DNNs. Through making deep hidden-to-hidden transitions, obvious performance improvements can be obtained, especially the LSTM-OP. Besides, the performance can also been improved by making deep input-to-hidden and hidden-to-output functions. It should be noted that, the LSTM-OP can yield comparable performance with the stacked LSTMs, which reached a similar conclusion with that in \cite{LSTMRNNAM}.

It is possible to design and train deeper variant of a LSTM network that combines different methods in Fig~\ref{Figure1} together. For instance, a stacked LSTM-OPs network may be constructed by combining the deep hidden-to-hidden transition and the stack of LSTMs. Combining different methods in Fig~\ref{Figure1} is a potential way to further improve the performance. Thus, experiments were conducted to evaluate some selected combinations of these methods for constructing deep LSTM networks, where each hidden layer had the same configuration as that in the experiments described above. The results are listed in Table~\ref{table2}, and the best performance can be obtained by combining the LSTM-OP and deep hidden-to-output function.

\begin{table} [t,h]
\caption{\label{table2} {Speech recognition results of selected combinations for constructing deep LSTM networks.}}
\vspace{2mm}
\centerline{
\begin{tabular}{|l|c|}
  \hline
  Model Descriptions & CER(\%) \\
  \hline \hline
  3-layer ReLU + LSTM-OP & 36.73 \\
  \hline
  2-layer Conv + 2-layer ReLU + LSTM-OP & 36.15 \\
  \hline
  LSTM-OP + 3-layer ReLU & \bf{34.65} \\
  \hline
  Stack of LSTM-IP (3-layer) & 35.00 \\
  \hline
  Stack of LSTM-OP (3-layer) & 34.84 \\
  \hline
\end{tabular}}
\end{table}

From these results in Table~\ref{table2}, we can find out that, the performance can be further improved by stacking LSTM-IPs and the LSTM-OPs. However, the network, that had LSTM-OP layer on top of three feed-forward intermediate layers, yielded worse performance than the LSTM-OP network, which needed to be further researched. What is noteworthy is that the network that had three full-connection hidden layers on top of LSTM-OP layer yielded the best performance, and required less computations than the stacked LSTM-OPs network in both training and testing procedures.

These experimental results had revealed that deep LSTM networks benefit from the depth. Compared with the shallow LSTM network, a 13.98\% relatively CER reduction can be obtained. Compared with the feed-forward DNNs, the deep LSTM networks can reduce the CER from 38.01\% to 34.65\%, which is a 8.87\% relatively CER reduction.

%Although deep LSTM networks have received lots of attentions, this paper studies the approaches of constructing deep LSTM networks from different perspectives, and suggests alternative architectures that can yield comparable performance with the LSTM projected and Stacked LSTMs proposed in the literatures \cite{SRWDRNN}\cite{HSRWDBLSTM}\cite{LSTMRNNAM}\cite{LSTMLVSP}.

\section{Discussion and Conclusions}
%In this paper, we have explored approaches to construct long short-term memory (LSTM) based deep recurrent neural networks (RNNs). Inspired from the discussion about how to construct deep RNNs in \cite{HTCDRNN}, several alternative architectures were proposed for deep LSTM networks. We implemented the LSTM training on multi-GPU devices with truncated BPTT, and empirically evaluated these architectures on a large vocabulary speech recognition task. The experimental results revealed that those constructed deep LSTM networks outperformed the standard shallow LSTM networks and DNNs. Besides, the LSTM-OP followed with three feed-forward layers outperformed the stacked LSTM-OPs, achieving the state-of-the-art performance on this task.

%In this paper, we have explored approaches to construct long short-term memory (LSTM) based deep recurrent neural networks (DRNNs). Inspired from the discussion about how to construct DRNNs in \cite{HTCDRNN}, several alternative architectures were proposed for deep LSTM, and empirically evaluated these architectures on a large vocabulary Mandarin conversational telephone speech recognition task. The experimental results revealed that those constructed deep LSTM networks outperformed the standard shallow LSTM networks. Besides, the LSTM-OP followed with three feed-forward layers outperformed the stacked LSTM-OPs, achieving the state-of-the-art performance on this task. We also show for the first time the LSTM networks can be quickly trained on multi-GPU devices.

In this paper, we have explored novel approaches to construct long short-term memory (LSTM) based deep recurrent neural networks (RNNs). A number of theoretical results support that a deep, hierarchical model can be more efficient at representing some functions than a shallow one \cite{LearningDeepArchAI}. This paper is focused on constructing deep LSTM networks, which have been shown to give state-of-the-art performance for acoustic modeling on some speech recognition tasks. Inspired from the discussion about how to construct deep RNNs in \cite{HTCDRNN}, several alternative architectures were constructed for deep LSTM networks from three points: (1) input-to-hidden function, (2) hidden-to-hidden transition and (3) hidden-to-output function. Furthermore, in this paper, some deeper variants of LSTMs were also designed by combining different points.

In this work, these LSTM network training were implemented on multi-GPU devices, in which the truncated BPTT learning algorithm was adopted, and the experiments discovered that the LSTM RNNs can also be quickly trained on GPU devices.

We empirically evaluated various deep LSTM networks on a large vocabulary Mandarin Chinese conversational telephone speech recognition task. The experiments revealed that constructing deep LSTM architecture outperformed the standard shallow LSTM networks and DNNs. Besides, the LSTM-OP followed with three feed-forward intermediate layers outperformed the stacked LSTM-OPs.%, achieving the state-of-the-art performance on this task.

However, we believe that this work is just a preliminary study on how to construct deep LSTM networks. There are many efforts need to be done about the architectures of LSTM networks. Some other architectures will be explored and evaluated in our future work, such as a LSTM-IP network which has three non-linear activation projection layers, a stacked LSTMs network followed with multiple feed-forward intermediate layers, a LSTM network with both input and output project layers, and deep architectures with the maxout unit improved LSTM layer \cite{MaxoutLSTM}.

\section{Acknowledgements}
\label{sec:Acknowledgement}
The work was supported in part by the National Basic Research Program (2013CB329304), the research special fund for public welfare industry of health(201202001), and National Natural Science Foundation (No.61121002, No.91120001).

%\vfill\pagebreak
\ninept
\bibliographystyle{IEEEbib}
\bibliography{strings,refs}

\end{document}